\documentclass[11pt,a4paper]{article}
\usepackage[hyperref]{emnlp2018}
\usepackage{times}
\usepackage{latexsym}

\usepackage{url}

\aclfinalcopy 

\usepackage[colorinlistoftodos,prependcaption]{todonotes}

\usepackage[utf8]{inputenc}
\usepackage[russian, arabic, english]{babel}

\usepackage{pgfplots}

\usepackage{dblfloatfix}

\usepackage{hyperref}
\hypersetup{
    colorlinks=true,
    linkcolor=blue,
    filecolor=magenta,      
    urlcolor=cyan,
}

\usepackage{subcaption}
\usepackage{pifont}

\usepackage{cleveref}

\usepackage{makecell}

\title{Filling Gender \& Number Gaps in Neural Machine Translation\\with Black-box Context Injection}
    
\author{Amit Moryossef$^\dagger$, Roee Aharoni$^\dagger$, Yoav Goldberg$^\dagger$$^*$ \\
\texttt{\{first.last\}@gmail.com} \\
\\ $^\dagger$Bar Ilan University, Ramat Gan, Israel \\
$^*$Allen Institute for Artificial Intelligence
}

\date{}

\begin{document}
    \maketitle
    \begin{abstract}
        When translating from a language that does not morphologically mark information such as gender and number into a language that does, translation systems must ``guess'' this missing information, often leading to incorrect translations in the given context. We propose a black-box approach for injecting the missing information to a pre-trained neural machine translation system, allowing to control the morphological variations in the generated translations \textit{without} changing the underlying model or training data. We evaluate our method on an English to Hebrew translation task, and show that it is effective in injecting the gender and number information and that supplying the correct information improves the translation accuracy in up to 2.3 BLEU on a female-speaker test set for a state-of-the-art online black-box system. Finally, we perform a fine-grained syntactic analysis of the generated translations that shows the effectiveness of our method. 
    \end{abstract}
    
    \section{Introduction}
    
    A common way 
    for marking information about gender, number, and case in language is morphology, or the structure of a given word in the language. However, different languages mark such information in different ways -- for example, in some languages gender may be marked on the head word of a syntactic dependency relation, while in other languages it is marked on the dependent, on both, or on none of them \cite{10.2307/415601}.
    This morphological diversity creates a challenge for machine translation, as there are ambiguous cases where more than one correct translation exists for the same source sentence. For example, while the English sentence ``I love language'' is ambiguous with respect to the gender of the speaker, Hebrew marks verbs for the gender of their subject and does not allow gender-neutral translation. This allows two possible Hebrew translations -- one in a masculine and the other in a feminine form. 
    As a consequence, a sentence-level translator (either human or machine) must commit to the gender of the speaker, adding information that is not present in the source. Without additional context, this choice must be done arbitrarily by relying on language conventions, world knowledge or statistical (stereotypical) knowledge. 
    
    Indeed, the English sentence ``I work as a doctor'' is translated into Hebrew by Google Translate using the masculine verb form \emph{oved}, indicating a male speaker, while ``I work as a nurse'' is translated with the feminine form \emph{ovedet}, indicating a female speaker (verified on March 2019). While this is still an issue, there have been recent efforts to reduce it for specific language pairs.\footnote{\url{blog.google/products/translate/reducing-gender-bias-google-translate/}}
    
    
    We present a simple black-box method to influence the interpretation chosen by an NMT system in these ambiguous cases. More concretely, we construct pre-defined textual hints about the gender and number of the speaker and the audience (the interlocutors), which we concatenate to a given input sentence that we would like to translate accordingly. We then show that a black-box NMT system makes the desired morphological decisions according to the given hint, even when no other evidence is available on the source side. While adding those hints results in additional text on the target side, we show that it is simple to remove, leaving only the desired translation.
    
    Our method is appealing as it only requires simple pre-and-post processing of the inputs and outputs, without considering the system internals, or requiring specific annotated data and training procedure as in previous work \cite{vanmassenhove-hardmeier-way:2018:EMNLP}. We show that in spite of its simplicity, it is effective in resolving many of the ambiguities and improves the translation quality in up to 2.3 BLEU when given the correct hints, which may be inferred from text metadata or other sources. Finally, we perform a fine-grained syntactic analysis of the translations generated using our method which shows its effectiveness. 
    
    \section{Morphological Ambiguity in Translation}

    Different languages use different morphological features marking different properties on different elements. For example, English marks for number, case, aspect, tense, person, and degree of comparison. However, English does not mark gender on nouns and verbs. Even when a certain property is marked, languages differ in the form and location of the marking \cite{10.2307/415601}. For example, marking can occur on the head of a syntactic dependency construction, on its argument, on both (requiring \emph{agreement}), or on none of them. 
    Translation systems must generate correct target-language morphology as part of the translation process. This requires knowledge of both the source-side and target-side morphology. Current state-of-the-art translation systems do capture many aspects of natural language, including morphology, when a relevant context is available \cite{Dalvi2017UnderstandingAI,bawden2018evaluating}, but resort to ``guessing'' based on the training-data statistics when it is not. Complications arise when different languages convey different kinds of information in their morphological systems. In such cases, a translation system may be required to remove information available in the source sentence, or to add information not available in it, where the latter can be especially tricky.

    \section{Black-Box Knowledge Injection}\label{sec:control}
    
    
    Our goal is to supply an NMT system with knowledge regarding the speaker and interlocutor of first-person sentences, in order to produce the desired target-side morphology when the information is not available in the source sentence. The approach we take in the current work is that of \emph{black-box injection}, in which we attempt to inject knowledge to the input in order to influence the output of a trained NMT system, without having access to its internals or its training procedure as proposed by \newcite{vanmassenhove-hardmeier-way:2018:EMNLP}.
    
    We are motivated by recent work by \citet{voita-EtAl:2018:Long} who showed that NMT systems learn to track coreference chains when presented with sufficient discourse context. We conjecture that there are enough sentence-internal pronominal coreference chains appearing in the training data of large-scale NMT systems, such that state-of-the-art NMT systems can and do track sentence-internal coreference. We devise a wrapper method to make use of this coreference tracking ability by introducing artificial antecedents that unambiguously convey the desired gender and number properties of the speaker and audience.
    
    More concretely, a sentence such as ``I love you'' is ambiguous with respect to the gender of the speaker and the gender and number of the audience. However, sentences such as ``I love you, she told him'' are unambiguous given the coreference groups \{I, she\} and \{you, him\} which determine \emph{I} to be feminine singular and \emph{you} to be masculine singular. We can thus inject the desired information by prefixing a sentence with short generic sentence fragment such as ``She told him:'' or ``She told them that'', relying on the NMT system's coreference tracking abilities to trigger the correctly marked translation, and then remove the redundant translated prefix from the generated target sentence. We observed that using a parataxis construction (i.e. ``she said to him:'') almost exclusively results in target-side parataxis as well (in 99.8\% of our examples), making it easy to identify and strip the translated version from the target side. Moreover, because the parataxis construction is grammatically isolated from the rest of the sentence, it can be stripped without requiring additional changes or modification to the rest of the sentence, ensuring grammaticality.
    \section{Experiments \& Results}\label{sec:experiments}
    
    
    
    To demonstrate our method in a black-box setting, we focus our experiments on Google's machine translation system (GMT), accessed through its Cloud API. To test the method on real-world sentences, we consider a monologue from the stand-up comedy show ``Sarah Silverman: A Speck of Dust''. The monologue consists of 1,244 English sentences, all by a female speaker conveyed to a plural, gender-neutral audience. Our parallel corpora consists of the 1,244 English sentences from the transcript, and their corresponding Hebrew translations based on the Hebrew subtitles.\footnote{The data is obtained from \url{www.opensubtitles.org}} We translate the monologue one sentence at a time through the Google Cloud API. Eyeballing the results suggest that most of the translations use the incorrect, but default, masculine and singular forms for the speaker and the audience, respectively. We expect that by adding the relevant condition of ``female speaking to an audience'' we will get better translations, affecting both the gender of the speaker and the number of the audience.
    
    
    \begin{figure*}[!ht]
        \begin{minipage}[t]{0.49\linewidth}
        \includegraphics[width=\linewidth]{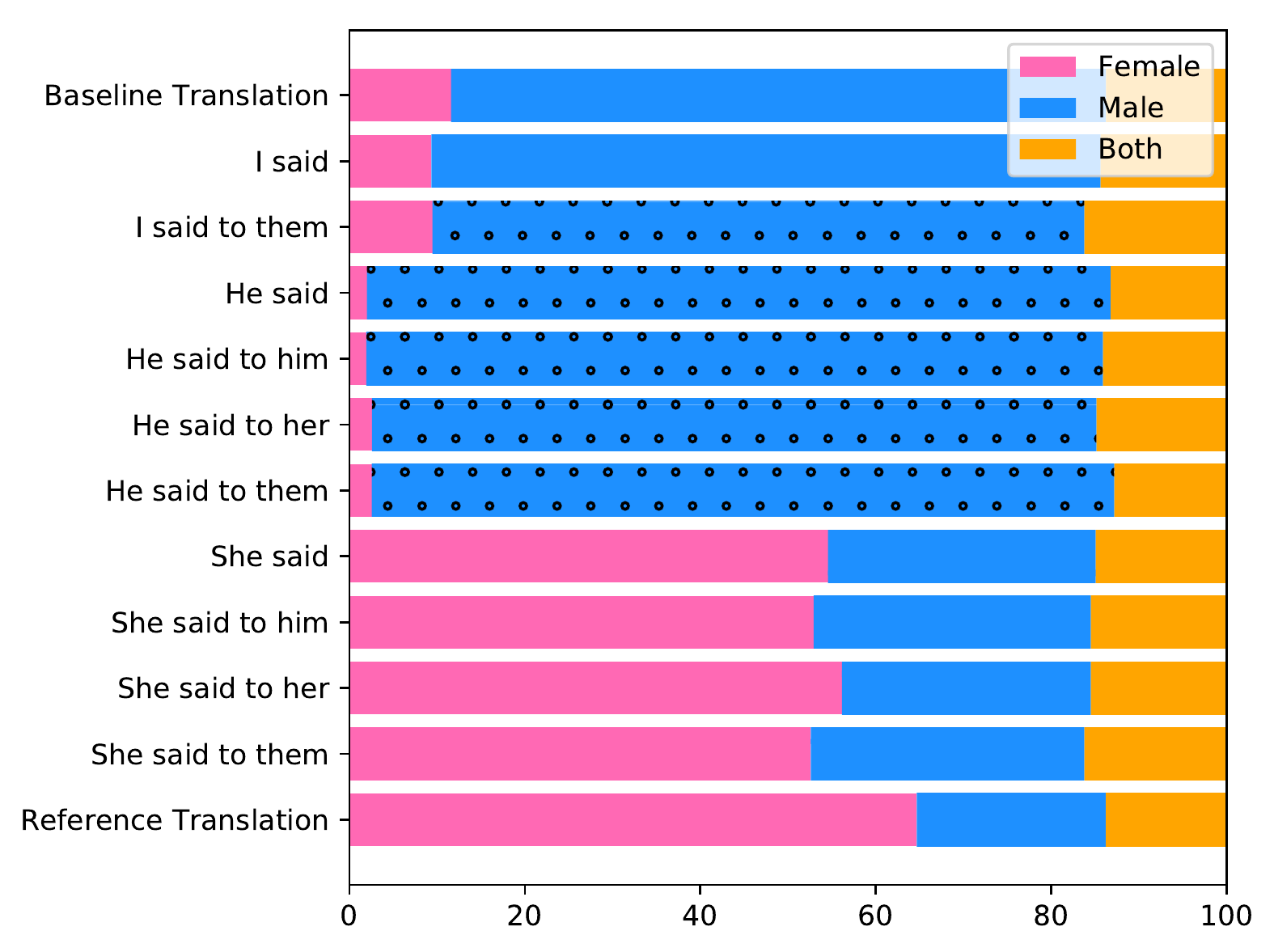}
            \caption{Gender inflection statistics for verbs governed by first-person pronouns.}
            \label{fig:speaker-gender-verb}
        \label{fig:rolling}
        \end{minipage}\hfill
        \begin{minipage}[t]{0.49\linewidth}
            \includegraphics[width=\linewidth]{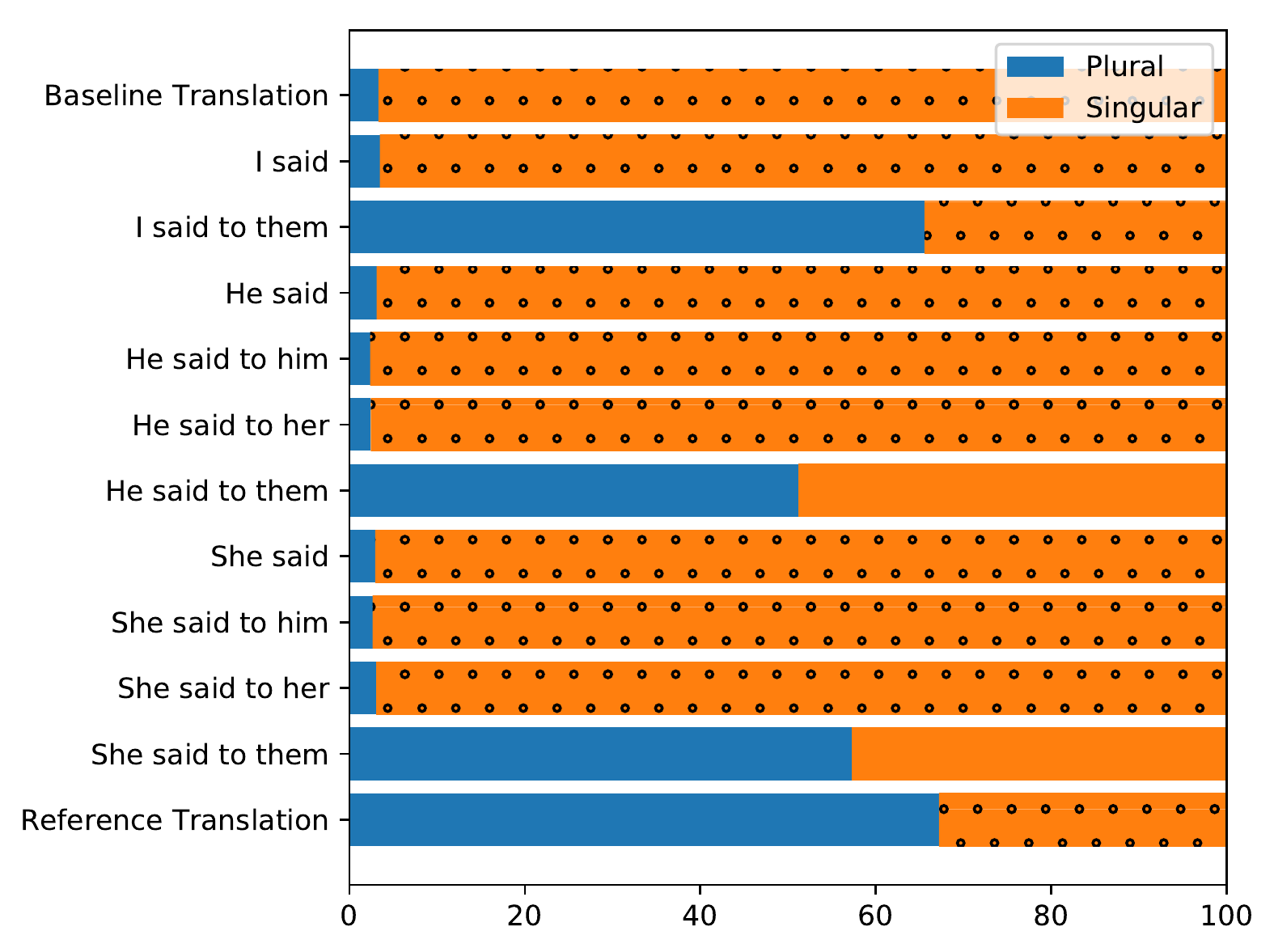}
            \caption{Number inflection statistics for second-person pronouns.}
            \label{fig:interlocutor-number}
        \end{minipage}
    \end{figure*}
    
    \begin{table}[t]


\centering
\begin{tabular}{|l|l|l|}
\hline
Speaker & Audience & BLEU \\
\Xhline{2\arrayrulewidth}
\multicolumn{2}{|c|}{Baseline}& 18.67 \\ \Xhline{2\arrayrulewidth}

He  & --    &       19.2    \\ \hline
He  & him   &       19.25   \\ \hline 
He  & her   &       19.3    \\ \hline 
He  & them  &       19.5    \\ \hline
I   & --    &       19.84   \\ \hline 
I   & them  &       20.23   \\ \hline 
She & --    &       20.8    \\ \hline 
She & him   &       20.82   \\ \hline 
She & her   &       20.98   \\ \hline
She & them  &       20.97   \\ \hline 

            \end{tabular}
            \caption{BLEU results on the Silverman dataset}
            \label{table:results-silverman}
            
    \end{table}
    
    To verify this, we experiment with translating the sentences with the following variations:
        \textbf{No Prefix}---The baseline translation as returned by the
        GMT system.
        \textbf{``He said:''}---Signaling a male speaker. We expect to further skew the system towards masculine forms.
        \textbf{``She said:''}---Signaling a female speaker and unknown audience. As this matches the actual speaker's gender, we expect an improvement in translation of first-person pronouns and verbs with first-person pronouns as subjects.
        \textbf{``I said to them:''}---Signaling an unknown speaker and plural audience.
        \textbf{``He said to them:''}---Masculine speaker and plural audience. 
        \textbf{``She said to them:''}---Female speaker and plural audience---the complete, correct condition. We expect the best translation accuracy on this setup.
        \textbf{``He/she said to him/her''}---Here we set an (incorrect) singular gender-marked audience, to investigate our ability to control the audience morphology.
    
    
    \subsection{Quantitative Results}\label{sec:evaluation}
    We compare the different conditions by comparing BLEU \cite{papineni2002bleu} with respect to the reference Hebrew translations. We use the \texttt{multi-bleu.perl} script from the Moses toolkit \cite{koehn2007moses}.
    Table \ref{table:results-silverman} shows BLEU scores for the different prefixes. The numbers match our expectations: Generally, providing an incorrect speaker and/or audience information decreases the BLEU scores, while providing the correct information substantially improves it - we see an increase of up to 2.3 BLEU over the baseline. 
    We note the BLEU score improves in all cases, even when given the wrong gender of either the speaker or the audience. We hypothesise this improvement stems from the addition of the word ``said'' which hints the model to generate a more ``spoken'' language which matches the tested scenario.
    Providing correct information for both speaker and audience usually helps more than providing correct information to either one of them individually. 
    The one outlier is providing ``She'' for the speaker and ``her'' for the audience. While this is not the correct scenario, we hypothesise it gives an improvement in BLEU as it further reinforces the female gender in the sentence.
    
    \subsection{Qualitative Results}\label{sec:finer-grained}
    The BLEU score is an indication of how close the automated translation is to the reference translation, but does not tell us what exactly changed concerning the gender and number properties we attempt to control. We perform a finer-grained analysis focusing on the relation between the injected speaker and audience information, and the morphological realizations of the corresponding elements. We parse the translations and the references using a Hebrew dependency parser.\footnote{\href{https://www.cs.bgu.ac.il/~yoavg/software/hebparsers/hebdepparser/}{ https://www.cs.bgu.ac.il/\textasciitilde{}yoavg/software/hebparsers/ hebdepparser/}} In addition to the parse structure, the parser also performs morphological analysis and tagging of the individual tokens. We then perform the following analysis.
   
    \textbf{Speaker's Gender Effects:} We search for first-person singular pronouns with subject case (\emph{ani}, unmarked for gender, corresponding to the English \emph{I}), and consider the gender of its governing verb (or adjectives in copular constructions such as `I am nice'). The possible genders are `masculine', `feminine' and `both', where the latter indicates a case where the none-diacriticized written form admits both a masculine and a feminine reading. We expect the gender to match the ones requested in the prefix.
    
    \textbf{Interlocutors' Gender and Number Effects:} We search for second-person pronouns and consider their gender and number. For pronouns in subject position, we also consider the gender and number of their governing verbs (or adjectives in copular constructions). For a singular audience, we expect the gender and number to match the requested ones. For a plural audience, we expect the masculine-plural forms.
    
    
    \textbf{Results: Speaker.} Figure \ref{fig:speaker-gender-verb} shows the result for controlling the morphological properties of the speaker (\emph{\{he, she, I\} said}). It shows the proportion of gender-inflected verbs for the various conditions and the reference. We see that the baseline system severely under-predicts the feminine form of verbs as compared to the reference. The ``He said'' conditions further decreases the number of feminine verbs, while the ``I said'' conditions bring it back to the baseline level. Finally, the ``She said'' prefixes substantially increase the number of feminine-marked verbs, bringing the proportion much closer to that of the reference (though still under-predicting some of the feminine cases).
    
    \textbf{Results: Audience.} The chart in Figure \ref{fig:interlocutor-number} shows the results for controlling the number of the audience (\emph{...to them} vs nothing). It shows the proportion of singular vs. plural second-person pronouns on the various conditions. It shows a similar trend: the baseline system severely under-predicts the plural forms with respect to the reference translation, while adding the ``to them'' condition brings the proportion much closer to that of the reference.

    \subsection{Comparison to \newcite{vanmassenhove-hardmeier-way:2018:EMNLP}}
    Closely related to our work, \newcite{vanmassenhove-hardmeier-way:2018:EMNLP} proposed a method and an English-French test set to evaluate gender-aware translation, based on the Europarl corpus \cite{koehn2005europarl}. We evaluate our method (using Google Translate and the given prefixes) on their test set to see whether it is applicable to another language pair and domain. Table \ref{table:french} shows the results of our approach vs. their published results and the Google Translate baseline. As may be expected, Google Translate outperforms their system as it is trained on a different corpus and may use more complex machine translation models. Using our method improves the BLEU score even further.

    \begin{table}[h]
\begin{center}
    
\begin{tabular}{|l|l|l|}
\hline
                                     & Male  & Female \\ \hline
VHW \citeyearpar{vanmassenhove-hardmeier-way:2018:EMNLP} Baseline & 37.58 & 37.75  \\ \hline
VHW \citeyearpar{vanmassenhove-hardmeier-way:2018:EMNLP} + TAG    & 38.71 & 38.97  \\ \hline
Google Translate                     & 39.33 & 39.02  \\ \hline
Google Translate + Prefix            & 39.95 & 39.95  \\ \hline
\end{tabular}
\caption{Comparison of our approach (using Google Translate) to \citet{vanmassenhove-hardmeier-way:2018:EMNLP} on their English-French gender corpus.}
\label{table:french}

\end{center}
\end{table}
    
    \newcommand\good[1]{{\color[HTML]{32cb00} #1}}
\newcommand\bad[1]{{\color[HTML]{FD6864} #1}}

\begin{table*}[t]
	\centering
\begin{tabular}{|l|l|l|l|}
\hline
           & English Text          & Masculine                                    & Feminine                                     \\ \hline
Hebrew     & I am nice             & ani nehmad                                   & ani nehmada                                  \\ \hline
\multicolumn{2}{|l|}{Prefix}       & ani \good{nehmad}           & ani \good{nehmada}          \\ \hline
Spanish    & I am delighted        & Estoy encantado                              & Estoy encantada                              \\ \hline
\multicolumn{2}{|l|}{Prefix}       & Estoy \good{encantado}      & Estoy \good{encantada}      \\ \hline
Portuguese & I was called          & Eu fui chamado                               & Eu fui chamada                               \\ \hline
\multicolumn{2}{|l|}{Prefix}       & Eu fui \good{chamado}       & Eu fui \bad{chamado}       \\ \hline
French     & I am patient          & je suis patient                              & je suis patiente                             \\ \hline
\multicolumn{2}{|l|}{Prefix}       & je suis \good{patient}      & je suis \good{patiente}     \\ \hline
Italian    & I am beautiful        & Sono bello                                   & Sono bella                                   \\ \hline
\multicolumn{2}{|l|}{Prefix}       & io sono \good{bello}        & io sono \good{bella}        \\ \hline
Russian    & I wrote a message     & \foreignlanguage{russian}{Я написал сообщение}& \foreignlanguage{russian}{Я написала сообщение} \\ \hline
\multicolumn{2}{|l|}{Prefix}       & \foreignlanguage{russian}{Я \good{написал} сообщение}  & \foreignlanguage{russian}{Я \bad{написал} сообщение}  \\ \hline
Czech      & I gave her the flower & já jsem ji dal květinu                       & já jsem ji dala květinu                      \\ \hline
\multicolumn{2}{|l|}{Prefix}       & \bad{Dala} jsem jí květinu & \good{Dala} jsem jí květinu \\ \hline
Romanian   & I am patient          & Sunt răbdător                                & Sunt răbdătoare                              \\ \hline
\multicolumn{2}{|l|}{Prefix}       & Sunt \good{răbdător}        & Sunt \good{răbdătoare}      \\ \hline
Catalan    & I am rich             & sóc ric                                      & sóc rica                                     \\ \hline
\multicolumn{2}{|l|}{Prefix}       & sóc \good{ric}              & sóc \bad{ric}              \\ \hline
Polish     & I am nice             & Jestem miły                                  & Jestem miła                                  \\ \hline
\multicolumn{2}{|l|}{Prefix}       & Jestem \good{miły}          & Jestem \good{miła}          \\ \hline
\end{tabular}
\caption{Examples of languages where the speaker's gender changes morphological markings in different languages, and translations using the prefix ``He said:'' or ``She said:'' accordingly}
\label{table:languages}
\end{table*}
    
    \subsection{Other Languages}
    To test our method’s outputs on multiple languages, we run our pre-and post-processing steps with Google Translate using examples we sourced from native speakers of different languages. For every example we have an English sentence and two translations in the corresponding language, one in masculine and one in feminine form.
    Not all examples are using the same source English sentence as different languages mark different information.
    Table \ref{table:languages} shows that for these specific examples our method worked on $6/10$ of the languages we had examples for, while for $3/10$ languages both translations are masculine, and for $1$ language both are feminine.

    \section{Related Work}\label{sec:related-work}
    \newcite{E17-1101} showed that given input with author traits like gender, it is possible to retain those traits in Statistical Machine Translation (SMT) models.
    \newcite{W17-4727} showed that incorporating morphological analysis in the decoder improves NMT performance for morphologically rich languages. 
    \newcite{burlot:hal-01618387} presented a new protocol for evaluating the morphological competence of MT systems, indicating that current translation systems only manage to capture some morphological phenomena correctly.
    Regarding the application of constraints in NMT, \newcite{N16-1005} presented a method for controlling the politeness level in the generated output. \newcite{DBLP:journals/corr/FiclerG17aa} showed how to guide a neural text generation system towards style and content parameters like the level of professionalism, subjective/objective, sentiment and others. 
    \newcite{W17-4811} showed that incorporating more context when translating subtitles can improve the coherence of the generated translations.
    Most closely to our work, \newcite{vanmassenhove-hardmeier-way:2018:EMNLP} also addressed the missing gender information by training proprietary models with a gender-indicating-prefix. We differ from this work by treating the problem in a black-box manner, and by addressing additional information like the number of the speaker and the gender and number of the audience. 
    \section{Conclusions}
    We highlight the problem of translating between languages with different morphological systems, in which the target translation must contain gender and number information that is not available in the source. We propose a method for injecting such information into a pre-trained NMT model in a black-box setting. We demonstrate the effectiveness of this method by showing an improvement of 2.3 BLEU in an English-to-Hebrew translation setting where the speaker and audience gender can be inferred. We also perform a fine-grained syntactic analysis that shows how our method enables to control the morphological realization of first and second-person pronouns, together with verbs and adjectives related to them. In future work we would like to explore automatic generation of the injected context, or the use of cross-sentence context to infer the injected information.
    
    \bibliography{emnlp2018}
    \bibliographystyle{acl_natbib_nourl}
    
\end{document}